\documentclass[sigconf,nonacm]{aamas} 

\usepackage{balance} 
\usepackage{algpseudocode}
\usepackage{algorithm}
\usepackage{tikz}
\usepackage{colortbl}

\acmSubmissionID{365}

\title[AAMAS-2023 RP-PU-MAPF]{Online Re-Planning and Adaptive Parameter Update for Multi-Agent Path Finding with Stochastic Travel Times}

\author{Atsuyoshi Kita}
\affiliation{
  \institution{Panasonic Holdings Corporation}
  \city{Osaka}
  \country{Japan}}
\email{kita.atsuyoshi@jp.panasonic.com}

\author{Nobuhiro Suenari}
\affiliation{
  \institution{Panasonic Holdings Corporation}
  \city{Osaka}
  \country{Japan}}
\email{suenari.nobuhiro@jp.panasonic.com}

\author{Masashi Okada}
\affiliation{
  \institution{Panasonic Holdings Corporation}
  \city{Osaka}
  \country{Japan}}
\email{okada.masashi001@jp.panasonic.com}

\author{Tadahiro Taniguchi}
\affiliation{
  \institution{Ritsumeikan University and Panasonic Holdings Corporation}
  \city{Shiga}
  \country{Japan}}
\email{taniguchi@ci.ritsumei.ac.jp}

\begin{abstract}
This study explores the problem of Multi-Agent Path Finding with continuous and stochastic travel times whose probability distribution is unknown. Our purpose is to manage a group of automated robots that provide package delivery services in a building where pedestrians and a wide variety of robots coexist, such as delivery services in office buildings, hospitals, and apartments. It is often the case with these real-world applications that the time required for the robots to traverse a corridor takes a continuous value and is randomly distributed, and the prior knowledge of the probability distribution of the travel time is limited. Multi-Agent Path Finding has been widely studied and applied to robot management systems; however, automating the robot operation in such environments remains difficult. We propose 1) online re-planning to update the action plan of robots while it is executed, and 2) parameter update to estimate the probability distribution of travel time using Bayesian inference as the delay is observed. We use a greedy heuristic to obtain solutions in a limited computation time. Through simulations, we empirically compare the performance of our method to those of existing methods in terms of the conflict probability and the actual travel time of robots. The simulation results indicate that the proposed method can find travel paths with at least 50\% fewer conflicts and a shorter actual total travel time than existing methods.
The proposed method requires a small number of trials to achieve the performance because the parameter update is prioritized on the important edges for path planning, thereby satisfying the requirements of quick implementation of robust planning of automated delivery services. 
\end{abstract}


\newcommand{\BibTeX}{\rm B\kern-.05em{\sc i\kern-.025ema  b}\kern-.08em\TeX}

\begin{document}


\pagestyle{fancy}
\fancyhead{}


\maketitle 


\section{Introduction}
\begin{figure*}[tb]
  \centering
  \includegraphics[width=0.75\linewidth]{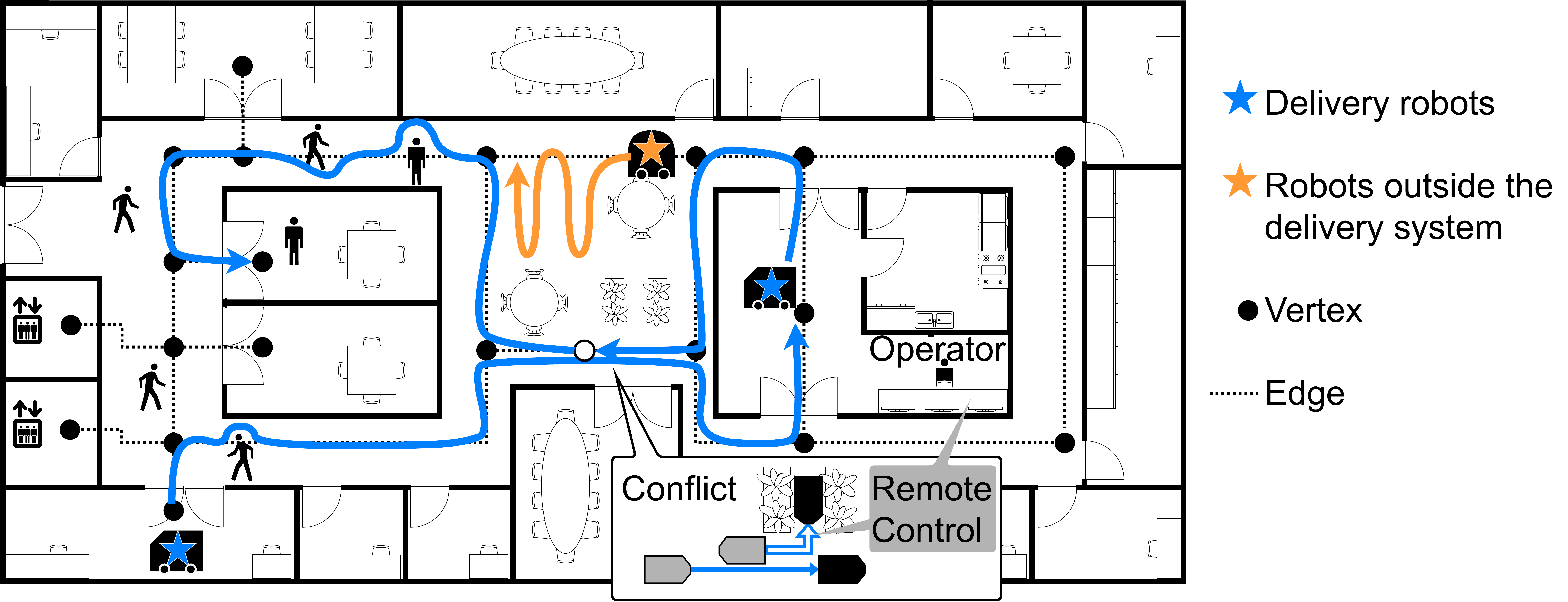}
  \caption{Overview of the proposed operation system for robot delivery service. 
  Automated robots ({\color[HTML]{007FFF}{$\bigstar$}}) deliver products from the store to customers in conference rooms in the building, 
  while a remote operator monitors multiple robots. 
  The robots proceed by avoiding pedestrians and cleaning robots ({\color[HTML]{FF9933}{$\bigstar$}}) that are directed by other systems. 
  If a conflict occurs due to an unexpected delay (as shown at vertex \textbigcircle), the remote operator manually operates the robot to retreat to a nearby retreat area (in this case, between plantings) and allow another robot to pass.}
  \label{fig:multipul_robots}
\end{figure*}

Expectations for delivery services using automated robots 
are rising because of the increasing demand for home deliveries and the labor
shortages in the delivery field in recent years. 
This study focuses on delivery services in buildings, 
such as office buildings, hospitals, and apartments, as shown in Figure \ref{fig:multipul_robots}. 
Building aisles are narrow; hence robots in automatic operation mode cannot pass by each other. 
Also, in these aisles are moving obstacles, such as people passing by or other groups of automated robots, 
which are under the control of another system. 
Robots can pass by pedestrians but have to slow down or temporarily stop. 
In a corridor with many people passing by, passing through takes longer than originally expected. 
Consequently, the time it requires to traverse a corridor is stochastically distributed. 
This is referred to as aleatoric uncertainty. 
On the other hand, epistemic uncertainty is the usual lack of information on the probability distribution 
of the travel time of each corridor.
The purpose of this paper is to propose an efficient 
method of operating automated delivery robots in the corridor of buildings under
both aleatoric and epistemic uncertainty.

Multi-Agent Path Finding (MAPF) has been studied and applied to warehouses ~\cite{Wurman2008} and airport operations ~\cite{Morris2016}.
Conflict Based Search (CBS)\cite{Sharon2015} and its extensions have been presented to efficiently solve the MAPF problem (\cite{Gange2021}, \cite{LiHarabor2021}).
The problem has extended its application from grid graphs to real-world situations. 
The Continuous-CBS~\cite{Andreychuk2019} incorporate graphs with continuous travel time, while 
the Large Agent MAPF~\cite{LiSurynek2019} considers agents with a physical shape. Cohen et al.~\cite{Cohen2019} extended the problem to the task of motion planning. Liu et al.~\cite{LiuMa2019} extended it to pickup-delivery tasks. 

However, aleatoric and epistemic uncertainty of the traffic condition in buildings
makes it difficult to automate the robot operation system fully.
Recent studies on the \textit{p}-Robust CBS~\cite{Atzmon2020} and STT-CBS~\cite{Peltzer2020} 
successfully addressed the aleatoric uncertainty by modeling the delay of travel time using probability distributions. 
The STT-CBS models the delay using a gamma distribution and creates the travel paths of agents whose conflict probability is less than a certain value.
However, in practice, the parameters of the distribution are not obvious or unstable.
As a result, robots actually get stuck with each other in the corridor, 
requiring remote operators to monitor and resolve the collision.
As described in Figure \ref{fig:multipul_robots}, the remote operator moves the robot to a nearby retreat area 
(e.g., a hollow in the corridor or a small space between obstacles) to allow another robot to pass. 
Because the automated robot keeps a certain distance from obstacles and walls for safety reasons, it cannot enter the retreat area. 
However, the remote operator can manipulate the robot, albeit slowly, to the retreat area.
These monitoring and remote operating take a long time, making it difficult for a single remote operator to monitor multiple robots.

This paper mainly describes two methods of coping with such situations: 1) parameter update (PU): the parameters of probability distribution for the stochastic travel times are adaptively updated using the Bayesian inference instead of being predefined; and 2) online re-planning (OR): the travel paths of robots are re-calculated and updated using updated parameters.

We formulate an online version of MAPF for graphs with continuous and stochastic travel times and propose an algorithm that efficiently solves the problem by introducing online re-planning and parameter update to the STT-CBS. We also demonstrate the importance of the algorithm through simulations with explicit uncertainty assumptions.

\section{Related Work}
\subsection{Multi-Agent Coordination under Uncertainty}
Zhou and Tokekar~\cite{Zhou2021} reviewed recent studies on multi-agent coordination under uncertainty, failures, and adversarial attacks. Nam and Shell~\cite{Nam2017} studied a problem of multi-robot task allocation where the cost of each robot-task pairing is randomly distributed. Yang and Chakraborty~\cite{Yang2018} studied a problem of chance-constrained knapsack problem that the weights of the items are stochastic.

Several studies addressed MAPF under uncertainty. Most of them were based on grid graphs and modeled the agents to stay in the same cell instead of moving to the next cell at a certain delay probability, such as Ma et al.~\cite{Ma2017}. Wagner and Choset~\cite{WagnerChoset2017} proposed the UM* algorithm that explored the path in a joint belief space from the start belief state to the goal belief state. They improved the algorithm by using random restart (permuted UM*). 

Atzmon et al.~\cite{Atzmon2020} proposed the \textit{p}-Robust CBS algorithm that finds the shortest \textit{p}-robust path that can be executed without conflict with a probability greater than \textit{p}. They proposed two verifiers that calculate the global conflict probability of the path in a constraint tree node and concluded that the Monte-Calro verifier performs better. They used a ternary tree as the search tree of the CBS for the optimal algorithm and used a binary tree for the heuristic variation (Greedy \textit{p}-Robust CBS).

In contrast to the two abovementioned studies, Peltzer et al.~\cite{Peltzer2020} used non-unit time graphs and modeled agent delays at each node for a continuous positive value of time that follows a gamma distribution. They defined the conflict probability and computed it through a Monte-Carlo simulation during the CBS algorithm to find a solution such that the pairwise conflict probability of agents is less than a threshold value $\epsilon$.

All the abovementioned studies assumed that the uncertainty can be correctly modeled; therefore, aleatoric uncertainty is incorporated, 
whereas epistemic uncertainty is not.
However, in real situations, the information on delay time sometimes cannot be obtained before
the operation, or only a limited number of observations are available.
We approach this issue by updating the model parameter using the delay observed 
while executing the plan.

Although not directly applicable to our problem, in the context of the Multi-agent Markov decision process (MMDP) \cite{Boutilier1996}, especially when addressing the problem with model-based reinforcement learning (MBRL), it is necessary to deal with epistemic uncertainty in dynamics model learning. Studies on models that incorporate both aleatoric and epistemic uncertainty are being conducted to improve learning efficiency (~\cite{Chua2018}, ~\cite{Okada2020variational}, ~\cite{Okada2020planet}). 

\subsection{Online Re-planning of MAPF}
Several studies approached MAPF with online re-planning. {\v{S}}vancara et al.~\cite{Vsvancara2019} and Ho et al.~\cite{Ho2019} defined a problem type in which new agents 
appear during the plan execution at an unknown time. Ma ~\cite{Ma2021} theoretically analyzed the algorithms for this problem type from the perspective of which agents can update their current plan and of the quality of the updated plan.

Shahar et al. ~\cite{Shahar2021} considered a type of MAPF problem in which each move has upper and lower bounds of execution time. They then proposed algorithms that can find the optimal solution guaranteed to have no conflict for all values of execution time between bounds and online re-planning for the problem setting. Their experiment results showed that the solution cost can be reduced.

Levy et al. ~\cite{Levy2022} introduced uncertainty into MAPF by considering a situation where agents move differently from what was planned at a certain probability at every time step. They then proposed an online approach to solving the problem. Accordingly, they updated the plan when a potential conflict in the current plan was detected by re-calculating the paths of the agents involved in that potential conflict. Their empirical experiments showed that the online update effectively avoids conflicts.

Okumura et al. ~\cite{Okumura2021} addressed a type of MAPF in which delay occurs due to agents stagnating instead of advancing at each node with a certain probability by online and distributed planning.

The abovementioned studies suggested that the online re-planning is effective for MAPF with uncertainty. However, their results were limited to graphs on a grid; thus, applying them to real-world situations (e.g., continuously distributed travel time) is difficult. 
In this work, we propose an online re-planning approach for non-unit time graphs.


\section{Online Re-Planning and Parameter Update}

\subsection{Problem Setup}
\label {problem_setup}
\subsubsection{Stochastic Travel Time Model}
We consider a connected bidirectional graph $G=(V,E)$ consisting of a set $E$ of edges and a set $V$ of vertices. 
The default travel time $w(e)$ of edge $e$ is equal to the edge length.
We have $N$ agents, and agent $a_i \in \{1...N\}$ moves from a start vertex $start_i \in V$ to a goal vertex $goal_i \in V$.
Agent $a_i$ is assigned a path $p_i = \{c_1,\ldots,c_{n_i}\}$ consisting of $n_i$ commands $c_j = (u_j, v_j, d_j)$, which is either a \textit{move} command such that 
$$ u_j\ne v_j, (u_j, v_j)\in E, d_j=w(u_j, v_j)$$ 
or a \textit{wait} command such that 
$$ u_j = v_j, u_j\in V, d_j\in R^+.$$ 
A path is valid when $u^i_1 = start_i$, $v^i_{n_i} = goal_i$, and $v^i_{j-1} = u^i_j$ for $j \in {2...n_i}$.

Let us model a stochastic travel time using gamma distribution. Each edge $e \in E$ has a shape parameter $a(e)$ and a scale parameter $b(e)$. The actual travel time of the edge is $w(e) + x$ with $x\sim \mathrm{Gamma}(a(e), b(e))$.

\subsubsection{Traffic Rules and Conflict Resolution}
The traffic rules applying to graph $G$ are presented here. First, vertices $v \in G$ can be occupied by at most one agent. Second, no agent can enter an edge $e=(u, v) \in G$ when the edge in the opposite direction $(v, u)$ is occupied by one or more agents. Violations of the rules are considered as a conflict.
During the execution of the path, unexpected conflicts may occur due to random delays. 
These conflicts are resolved according to the following rules:
\begin{enumerate}
    \item Vertex conflict: If an agent $a_i$ finishes traversing an edge $(u, v)$ at time $t$, and another agent occupies the vertex $v$, agent $a_i$ stays at the edge and asks the remote operator for help. Agent $a_i$ is then removed from the edge $(u, v)$ at time $t+w((u, v))*C_{penalty}$, where $C_{penalty}\in R^+$ is a penalty constant. If no agents are traversing the edge $(u, v)$, the opposite edge $(v, u)$ becomes available at this moment. If agent $a_i$ has a next \textit{move} command $c=(u_{next}, v_{next}, d)$ to execute, the agent is inserted to the edge $(u_{next}, v_{next})$ at time $t+(w(u, v)+w(u_{next}, v_{next}))*C_{penalty}$ if the edge $(u_{next}, v_{next})$ is available. If the edge is unavailable at the moment, agent $a_i$ is inserted to the edge next time it becomes available.
    \item Edge conflict: When an agent $a_i$ tries to enter an edge $(u, v)$, and the edge in the opposite direction $(v, u)$ is occupied by one or more agents, agent $a_i$ has to wait until the edge $(v, u)$ becomes empty.
\end{enumerate}

\begin{figure}[tb]
  \centering
  \includegraphics[width=0.75\linewidth]{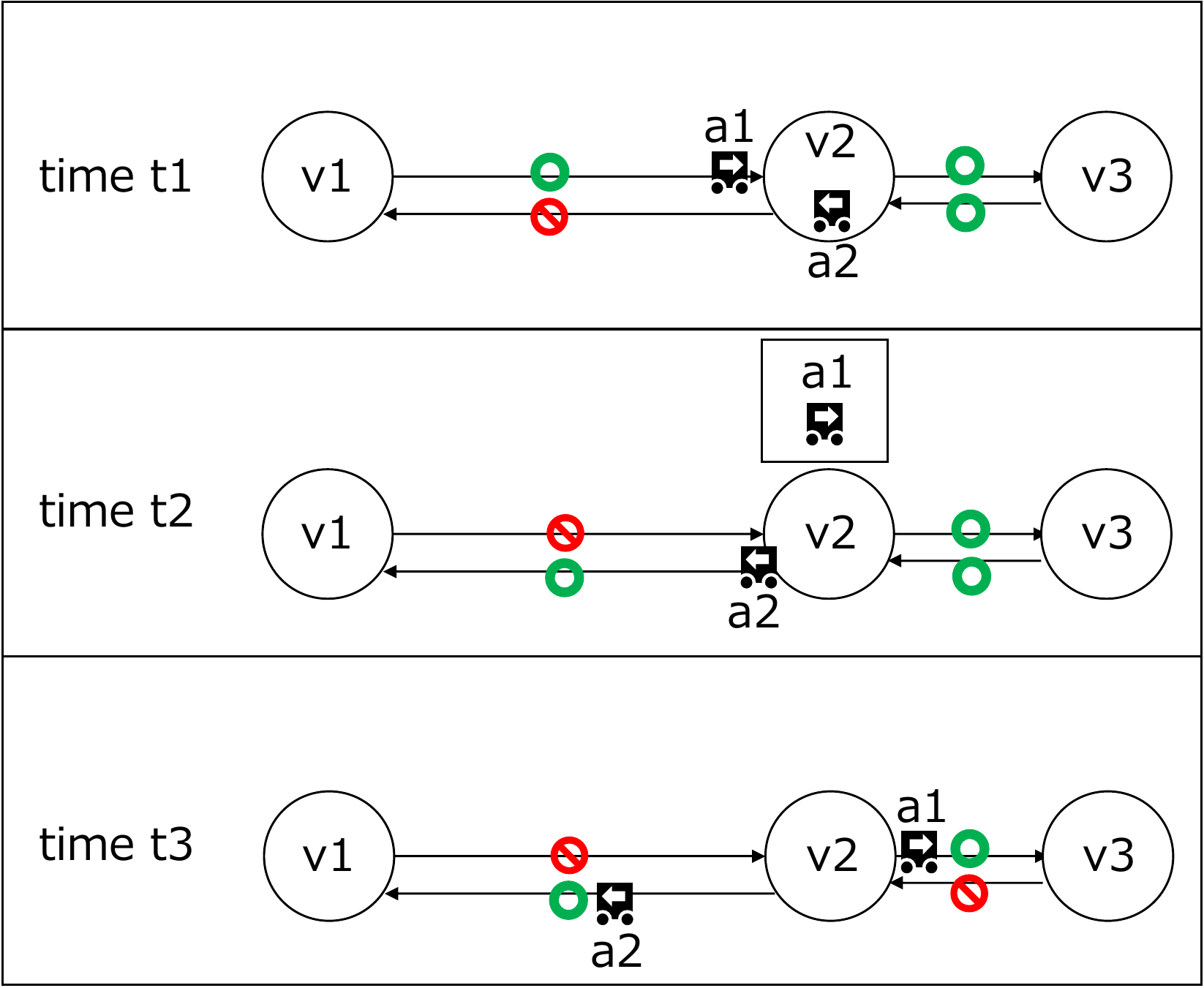}
  \caption{
  Sample operation for resolving the vertex conflict.
  At time $t_1$, agent $a_1$ tries to enter vertex $v_2$ occupied by agent $a_2$.
  At time $t_2=t_1+w((v_1, v_2))*C_{penalty}$, agent $a_1$ is removed from the edge $e(v_1, v_2)$. Edge $e(v_2, v_1)$ becomes available because no other agent is traversing the edge $e(v_1, v_2)$, and agent $a_2$ starts traversing $e(v_2, v_1)$. At time $t_3=t_2+w((v_2, v_3))*C_{penalty}$, agent $a_1$ is inserted to the edge $e(v_2, v_3)$, and the counter wise edge $e(v_3, v_2)$ becomes unavailable.
  }
  \label{fig:vertex_conflict}
\end{figure}

Figure \ref{fig:vertex_conflict} describes the vertex conflict resolution process. When the vertex conflict occurs, the remote operator detaches the stacked robot from the robot operation system and controls the robot manually. The remote operator moves the robot to a nearby waiting space (e.g., a hollow in the corridor) to allow another robot to pass. Because the automated robot keeps a certain distance from obstacles and walls for safety reasons, it cannot enter the waiting space. However, the remote operator can manipulate the robot, albeit slowly, to the waiting space. This slow manual operation is modeled with the penalty $C_{penalty}$.


\subsubsection{Online MAPF}
We consider a centralized robot operating system. The central controller has information about which command each agent is currently executing, and it can update the path of all agents at any time; however, the currently running command must remain unchanged, and other commands in the updated path must be consistent with it.

The offline MAPF problem instance for the offline solution consists of the graph $G$, $start_i$, and $goal_i$ for each agent $i\in{1\ldots N}$. The agents start at time $0$ from their start vertices. We now consider the problem instance for the online re-planning at time $t$. At this moment, agent $a_i$ is executing a \textit{move} or a \textit{wait} command $c_i(t) = (u_i(t), v_i(t), d_i(t))$. In the case of a \textit{move} command, the finish time of the command is uncertain because of the delay during the edge traversal, but the start time of the current command $st_i(t)$ and the planned finish time (the finish time assuming no delays) $ft_i(t) = st_i(t) + d_i(t) = st_i(t) + w(u_i(t), v_i(t))$ are available. We use these as the problem input. The \textit{fixed} command for agent $a_i$ is $(st_i(t), u_i(t), ft_i(t), v_i(t))$, and the path after the online update has to begin with it.

In the online MAPF problem, the problem instance at time $t$ consists of the following:
\begin{itemize}
    \item graph $G$;
    \item start time of the \textit{fixed} command $st_i(t)$;
    \item start vertex of the \textit{fixed} command $u_i(t)$;
    \item end time of the \textit{fixed} command $ft_i(t)$;
    \item end vertex of the \textit{fixed} command $v_i(t)$;
    \item goal vertex $goal_i$ for $i\in {1...N}.$; and
    \item calculation time limit $t_{limit}$.
\end{itemize}

The solution to the online MAPF problem comprises a list of commands for each agent $path_i = {c_1, c_2, \ldots, c_{n_i}}$, which begins from the \textit{fixed} command, such that $c_1=(u_1, v_1, d_1) = (u_i(t), v_i(t), ft_i(t)-st_i(t))$. The solution algorithm aims to find a solution in which the commands, including \textit{fixed} commands, can be executed without conflict. 

In the online re-planning context, the agents move while the central controller calculates the plan. If the states of an agent change before the calculation finishes, the new plan based on the old status may be invalid; therefore, the problem instance has a time limit for the calculation time. If the algorithm fails to return a solution within the time limit, the plan is not updated.

\subsection{Solution Algorithm}
Our algorithm for finding the updated paths for the online MAPF is based on the STT-CBS~\cite{Peltzer2020}. Algorithm \ref{alg:highlevel} underlines and presents the modifications described in this section.

\subsubsection{Greedy Heuristic}
In the best first search of the original STT-CBS, the node with the lowest cost is expanded. 
Instead, we introduce here the Greedy STT-CBS (GSTT-CBS) that selects the node with a min-max conflict probability. More precisely, for each constraint tree node with path $p_i= \{c_1,\ldots,c_{n_i}\}$ for agents $a_i\in\{1 \ldots N\}$, the maximum conflict probability of all command pairs in the solution 
$$P_{max} = \max_{c_k, c_l} \{P(c_k, c_l) : c_k \in p_i, c_l \in p_j, i\ne j, (i,j)\in\{1 \ldots N\}\}$$
are calculated through a Monte-Carlo simulation, where $P(c_k, c_l)$ is the conflict probability between commands $c_k$ and $c_l$. The node with a minimum value of $P_{max}$ is popped out of the priority queue and expanded.
The same heuristic is introduced into the Greedy p-Robust CBS\cite{Atzmon2020}. According to experiments, this is effective in reducing computation time.

The algorithm must return a solution within a calculation time limit; thus, the high level search returns the solution in the top of the priority queue, even if it still has a conflict when the algorithm reaches the calculation time limit. 

\subsubsection{Fixed Command}
The low level search is A* algorithm on a time-expanded graph $\mathcal{G} = (\mathcal{V}, \mathcal{E})$.
With the time horizon $T \in Z^+$, $\mathcal{V} = V \times \{0, 1, \ldots T\}$ and $\mathcal{E} = \{((v_1, \tau_1), (v_2, \tau_2)) \in \mathcal{V} \times \mathcal{V}: (v_1, v_2)\in E \wedge \tau_2-\tau_1 = w(v_1, v_2)\}$. Adding a constraint that prohibits traversing edge $(u, v)\in E$ from time $t$ to $t+w(u,v)$ corresponds to prohibiting (i.e., removing) the edge $((u, t),(v, t+w(u,v)))$ in $\mathcal{G}$.

Since the path for each agent have to begin with the \textit{fixed} command, the low level search finds the shortest path from the time expanded vertex $(v_i(t), ft_i(t))$ to one of the time expanded vertices $(goal_i, \tau)$ for $\tau \in {1...T}$, and then adds the \textit{fixed} command at the head of the path.
When a conflict is found in a CBS node, and the conflict is between a \textit{fixed} command and a non-\textit{fixed} command, instead of creating two child nodes, one child node with additional constraint forbidding the non-\textit{fixed} command is created and added to the search tree. 

\begin{algorithm}[tb]
  \caption{High Level Search}
  \label{alg:highlevel}
\begin{algorithmic}
\Function{find\_solution}{Online MAPF instance}
  \State Initialise root node $R$
  \State $R.constraints \gets \emptyset$
  \State $R.solution \gets$ Find low level solution to $R$ using A*
  \State $R.cost \gets$ Find solution cost
  \State Insert $R$ into Priority Queue
  \While{Priority Queue $\ne \emptyset$}
    \State $P \gets$ \underline{Pop the node with min-max conflict probability}
    \State $P.conflict \gets \mathrm{GetFirstConflict}(P)$ 
    \If{$P$ is conflict-free}
      \State \Return $P.solution$ 
    \ElsIf{\underline{timeout}}
      \State \underline{\Return $P.solution$}
    \EndIf
    \ForAll{$(a_i, c_i)$ involved in $P.conflict$}
      \If{\underline{command $c_i$ is fixed for agent $a_i$}}
        \State \underline{Continue}
      \EndIf
      \State Create child node $C$
      \State $C.const \gets P.const + \mathrm{Const}(P.conflict, a_i)$
      \State \underline{Find low level solution to $C$ using A*}
      \State $C.cost \ gets$ Find solution cost
      \State add $C$ to Priority Queue
    \EndFor
  \EndWhile
  \State \Return $x$
\EndFunction
\end{algorithmic}
\end{algorithm}

\subsection{Online Re-planning}
Various policies for online re-planning are possible regarding the time to re-plan and the selection of agents to re-plan. We use a constant interval policy, whereby all agent's plans are updated at regular intervals of time $t_{ci}$. Other policies can also be adopted (e.g., updating all agent's plans if the execution time of a command is delayed beyond a certain value or updating the plans of two agents when these agents are expected to conflict in the future).

\subsection{Parameter Update}
The STT-CBS algorithm uses a Monte-Carlo simulation to evaluate the conflict
probability between paths.
We assume that the true parameters of delay distribution, 
$a(e)$ and $b(e)$ for $e\in E$, are unknown. 
We only have prior knowledge, 
$a_{prior}(e)$ and $b_{prior}(e)$ for $e\in E$ 
and estimate the parameters from $m$ observations of delay $\mathbf{x}=(x_1, \ldots x_m)$ 
obtained as the agents traverse the edge $e$. We now drop the index $e$ for notation simplicity 
(e.g., $a=a(e) and b=b(e)$).



Let the prior distribution of $a$ and $b$ be
$$p(a,b \mid p,q,r,s)=K\frac{p^{a-1}\exp(-q/b)}{\Gamma(a)^r b^{as}}$$
with a normalization constant $K$ and parameters $p$, $q$, $r$, $s$. The 
posterior distribution is obtained as
\begin{eqnarray*}
P(a, b\mid \mathbf{x})  &\propto& \prod^m_{i=1} \mathrm{Gamma}(x_i|a,b) * p(a,b|p,q,r,s)\\
&=&  \frac{{p^{\prime}}^{a-1}\exp(-q^{\prime}/b)}{\Gamma(a)^{r^{\prime}} b^{as^{\prime}}},
\end{eqnarray*}
where $p^{\prime}=p \prod^m_{i=1} x_i$, $q^{\prime}=q + \sum^m_{i=1} x_i$, $r^{\prime}=r+m$, and $s^{\prime}=s+m$ are the parameters after the Bayesian update~\cite{Fink1997}. 
Differentiating the log posterior distribution and setting it to zero
$$\partial_a \ln P(a, b\mid \mathbf{x}) = \ln p^{\prime} - r^{\prime}\frac{\Gamma^{\prime}(a)}{\Gamma(a)} - s^{\prime} \ln b=0$$
$$\partial_b \ln P(a, b\mid \mathbf{x}) = \frac{q^{\prime}}{b^2} - \frac{as^{\prime}}{b} =0$$
yields simultaneous equations of the maximum a posteriori (MAP) estimator $a_{map}$ and $b_{map}$;
\begin{eqnarray*}
a_{map}  &=& \Psi^{-1}\left(\frac{\ln p^{\prime} - s^{\prime} \ln b_{map}}{r^{\prime}}\right),\\
b_{map}  &=& \frac{q^{\prime}}{a_{map} s^{\prime}}
\end{eqnarray*}
where $\Psi^{-1}$ is the inverse digamma function.
Eliminating $a_{map}$ from the equations derives the following;
$$
f(b) \coloneqq (\ln p^{\prime} - s^{\prime} \ln b) - r^{\prime} \Psi\left( \frac{q^{\prime}}{b s^{\prime}}\right).
$$
We use Newton's method 
$$
b \gets b - \frac{f(b)}{f^{\prime}(b)}
$$
to obtain the value of $b_{map}$, such that $f(b_{map}) = 0$.

The hyper parameters for prior distribution are $a_{prior}$, $b_{prior}$, $r$, and $s$.
The parameter $p$ and $q$ can be calculated as 
\begin{eqnarray*}
p &=& \exp\left(r\Psi(a_{prior}) - s\ln b_{prior}\right),\text{and}\\
q &=& a_{prior}b_{prior}s.
\end{eqnarray*}

\subsection{Overall Algorithm Flow}
Algorithm \ref{alg:overall} describes the overall flow of the parameter update and the online re-planning. We use a discrete event simulator to simulate the movement of agents with the random delay and traffic rules described in Section \ref{problem_setup}. The GSTT-CBS solver uses the gamma probability distribution model for the delay of each edge $e$ with prior of parameters $a_{prior}(e)$ and $b_{prior}(e)$ and MAP estimators $a_{map}(e)$ and $b_{map}(e)$. The simulator processes event (e.g., the arrival of an agent at a vertex or the entrance of an agent to an edge) one by one. Every time an agent enters an edge, a delay is randomly generated from the gamma distribution with true parameters $a(e)$ and $b(e)$. At each simulation step, the solver updates the model parameters $a_{map}(e)$ and $b_{map}(e)$ using the observed delays. When the re-plan policy is satisfied (i.e., when a certain time period has passed since the last update), the simulator converts the current state of the agents to an online MAPF problem instance, inputs it to the solver, and then updates the action plan of agents using the solution returned from the solver.

We consider a situation with a list of tasks (i.e., start and goal locations for each agent) for the same map and sequentially complete the tasks. The solver creates the plan for the next task using the parameters of the probability distribution updated with the delay time observed in the previous task. In short, the planning is based on a slightly more accurate model each time the agent passes an edge and obtains information.

\begin{algorithm}[tb]
  \caption{Overall Algorithm}
  \label{alg:overall}
\begin{algorithmic}
    \State Initialise Solver with prior knowledge
    \ForAll{problem instance $p_{ini}$ in problem list}
        \State Initialize the simulator Env with $p_{ini}$
        \State Env.PLAN $\gets$ solver.FIND\_SOLUTION($p_{ini}$)
        \While{not Env.IS\_DONE()}
            \State delay observation $obs \gets$ Env.STEP()
            \State Solver.UPDATE\_PARAMETER($obs$)
            \If{Env.REPLAN\_POLICY()}
                \State problem instance $p \gets$ Env.TO\_PROBLEM()
                \State Env.PLAN $\gets$ Solver.FIND\_SOLUTION($p$)
            \EndIf
        \EndWhile
    \EndFor
\end{algorithmic}
\end{algorithm}

\section{Experiment}
This section describes the design of the experiments and their results.
First, we evaluate our approach on non-grid graphs with uncertain stochastic travel times 
and compare the use of the greedy heuristic, the online re-planning (OR), and the parameter update (PU).
Next, we perform an analysis on the effect of each technique by using different parameter values or additional problem instances. 
The algorithms are implemented in C++ and run on a 2.7 GHz Intel Core i7-10850H laptop PC with 32 GB RAM.
\subsection{Experiment Design}

\subsubsection{Problem instance}
\begin{figure}[tb]
  \centering
  \includegraphics[width=1.0\linewidth]{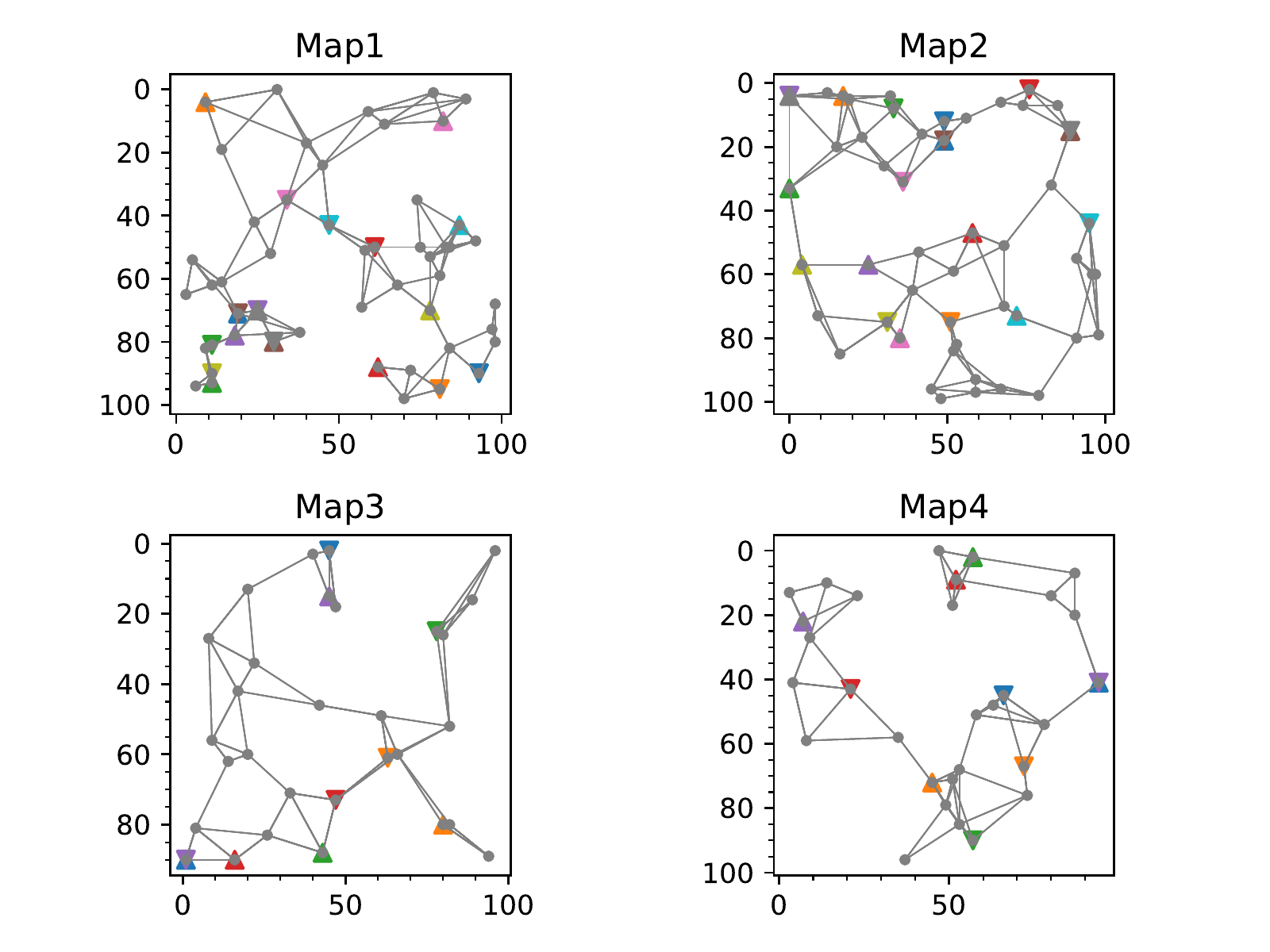}
  \caption{
  Graphs used for the experiment. Map1 and Map2 have 50 nodes and 10 agents. Map3 and Map4 have 30 nodes and 5 agents.
  The triangles are an example of start and goal vertices of agents.
  }
  \label{fig:problem_instance}
\end{figure}

We randomly generate non-grid graphs for the experiment. Vertices are generated with coordinate $(x, y)$ and degree $\delta$, such that
\begin{eqnarray*}
x &=& \mathrm{Random}(\{0, \ldots , 99\}),\\
y &=& \mathrm{Random}(\{0, \ldots , 99\}),\\
\delta &=& \mathrm{Random}(\{2, \ldots , 4\}).\\
\end{eqnarray*}
For each vertex $v$, edges are generated between $v$ and its $\delta$ nearest vertices. Figure \ref{fig:problem_instance} illustrates the generated graphs. 

The parameters of the delay probability distribution for the edge $e$ are randomly generated as follows:
\begin{itemize}
\item mean: $m_e = \mathrm{Random}(\{3, \ldots, 9\})*1.0$,
\item variance: $v_e = \mathrm{Random}(\{1, \ldots, 4\})*0.1$,
\item true shape parameter: $a(e) = m_e^2 v_e$,
\item true scale parameter: $b(e) = v_e/m_e$.
\end{itemize}

If the obtained graph is not connected, the same process is repeated until a connected graph is obtained. A total of  100 tasks are randomly generated for each map. The start and goal vertices are randomly selected from the vertices $V$, such that the agents do not share the start and goal vertices.

\subsubsection{Hyper Parameters}
The other hyper parameters are set as follows: The threshold of the pairwise conflict probability
between two agents $\epsilon$ in the STT-CBS and GSTT-CBS is 0.01; the time interval of the OR $t_{ci}$ is 100; the penalty constant for the conflict resolution by operator $C_{penalty}$ is set to 1.0; the calculation time limit of the initial planning and the online re-planning $t_{limit}$ is set to 10 seconds; and the parameters for prior distribution are $[a_{prior},b_{prior},r,s] = [1.0, 0.2, 0.1, 0.1]$. Setting $r$ and $s$ to 0.1 means that the parameter estimation is performed with more emphasis on observed values than on prior knowledge.

\subsection{Experiment Results}
Table \ref{mainresult} shows the average number of conflicts and the average of the flowtime over the execution of 100 tasks. Flowtime is defined as the sum of time the agents take to reach their goal vertex. The average calculation time in seconds for the initial solution and the ratio of the tasks that run out of time during the initial solution computation is also shown.

In terms of the number of conflicts and flowtime, the experimental result depicts that our proposed method outperforms the existing method (note the highlighted columns). 
All three methods (i.e., greedy heuristic, OR, and PU) were confirmed effective since the lack of any one of these methods resulted in lower performance.
In the next section, we will analyze the effect of each method in detail to answer the questions of how and why each of these methods is effective in reducing the number of conflicts and flowtime.

In terms of the calculation time and timeout ratio of initial solution, our proposed method is better than the existing method. The calculation time is short enough to meet the assumed system requirements.

\begin{table*}[bt]
    \centering
    \caption{Effect of greedy heuristic, online re-planning, and parameter update. Boldface and underlines indicate the best results. Underlines mean the second best.}
    {\small
    \begin{tabular}{lr|>{\columncolor[rgb]{0.86,0.86,0.86}}cccc|ccc>{\columncolor[rgb]{0.86,0.86,0.86}}c|cc}
    \toprule
        ~ & ~ & \multicolumn{4}{c|}{GSTT-CBS}  & \multicolumn{4}{c|}{STT-CBS}  & \multicolumn{2}{c}{CBS} \\ 
        ~ & ~ & with OR,PU & with PU & with OR & ~ & with OR,PU & with PU & with OR & ~ & with OR & ~ \\ 
        ~ & ~ & (ours) & ~ & ~ & ~ & ~ & ~ & ~ & \cite{Peltzer2020} & ~ & ~ \\ \hline
        \multicolumn{2}{l|}{Greedy Heuristic} & \checkmark & \checkmark & \checkmark & \checkmark & ~ & ~ & ~ & ~ & ~ & ~ \\ 
        \multicolumn{2}{l|}{Online Re-planning} & \checkmark & ~ & \checkmark & ~ & \checkmark & ~ & \checkmark & ~ & \checkmark & ~ \\ 
        \multicolumn{2}{l|}{Parameter Update} & \checkmark & \checkmark & ~ & ~ & \checkmark & \checkmark & ~ & ~ & ~ & ~ \\ \midrule
        Number of conflict & Map1 & \textbf{\underline{1.81}} & \underline{1.97} & 2.68 & 2.92 & 2.96 & 3.21 & 3.08 & 3.56 & 2.90 & 3.11 \\ 
        ~ & Map2 & \textbf{\underline{0.87}} & \underline{0.96} & 2.05 & 1.97 & 2.15 & 1.96 & 2.34 & 2.38 & 2.22 & 2.27 \\ 
        ~ & Map3 & \textbf{\underline{0.05}} & \textbf{\underline{0.05}} & 0.12 & 0.14 & 0.08 & 0.09 & 0.26 & 0.31 & 0.27 & 0.28 \\ 
        ~ & Map4 & \textbf{\underline{0.15}} & \underline{0.22} & 0.91 & 0.93 & 0.59 & 0.64 & 1.15 & 1.32 & 1.27 & 1.35 \\ \hline
        Flowtime & Map1 & \textbf{\underline{1334.91}} & \underline{1342.72} & 1383.96 & 1390.50 & 1418.99 & 1423.95 & 1408.13 & 1431.77 & 1388.45 & 1396.91 \\ 
        ~ & Map2 & \textbf{\underline{1175.94}} & \underline{1182.06} & 1273.48 & 1247.17 & 1274.46 & 1249.78 & 1287.13 & 1276.63 & 1279.06 & 1267.58 \\ 
        ~ & Map3 & \textbf{\underline{478.22}} & \underline{478.67} & 481.60 & 481.16 & 477.09 & 477.92 & 489.94 & 493.28 & 493.70 & 492.07 \\ 
        ~ & Map4 & \textbf{\underline{620.71}} & \underline{626.25} & 669.83 & 665.34 & 648.10 & 647.03 & 690.42 & 696.03 & 700.34 & 698.03 \\ \hline
        Calculation Time(sec) & Map1 & 3.70 & 3.67 & \textbf{\underline{3.04}} & \underline{3.07} & 6.23 & 6.26 & 5.05 & 5.10 & 4.03 & 4.04 \\ 
        ~ & Map2 & \textbf{\underline{2.65}} & \underline{2.71} & 3.08 & 3.04 & 5.14 & 5.02 & 4.50 & 4.42 & 4.06 & 4.07 \\ 
        ~ & Map3 & \textbf{\underline{0.17}} & \underline{0.18} & \underline{0.18} & 0.19 & 0.49 & 0.49 & 0.40 & 0.40 & 0.36 & 0.36 \\ 
        ~ & Map4 & \underline{0.66} & \textbf{\underline{0.50}} & 1.04 & 0.95 & 2.34 & 2.43 & 2.93 & 2.97 & 2.91 & 2.89 \\ \hline
        Timeout rate & Map1 & 0.27 & 0.24 & \textbf{\underline{0.22}} & \textbf{\underline{0.22}} & 0.54 & 0.54 & 0.44 & 0.44 & 0.34 & 0.34 \\ 
        ~ & Map2 & \textbf{\underline{0.16}} & \underline{0.17} & 0.21 & 0.20 & 0.44 & 0.41 & 0.39 & 0.39 & 0.35 & 0.35 \\ 
        ~ & Map3 & \textbf{\underline{0.01}} & \textbf{\underline{0.01}} & \textbf{\underline{0.01}} & \textbf{\underline{0.01}} & 0.03 & 0.03 & 0.03 & 0.03 & 0.03 & 0.03 \\ 
        ~ & Map4 & \underline{0.03} & \textbf{\underline{0.01}} & 0.07 & 0.06 & 0.17 & 0.17 & 0.24 & 0.26 & 0.26 & 0.26 \\ \bottomrule
    \end{tabular}
    }

    \label{mainresult}
\end{table*}

\subsection{Analysis}
\subsubsection{Effect of the greedy heuristic}
We examine here the simulation result on Map1 in Table \ref{mainresult} to clarify why the greedy heuristic is effective. We divide the 100 tasks into ``easy'' and ``difficult'' tasks.
The former comprises the tasks from which the vanilla CBS algorithm found a conflict-free solution within the time limit. 
The latter comprises the tasks in which the CBS algorithm reached a timeout. 
Table \ref{timeout_analysis} shows the performance of each algorithm for both groups.

STT-CBS fails to find a conflict-free solution for 10 tasks in the ``easy'' group.
This is because STT-CBS takes a longer time to detect conflict compared to CBS and can expand a lesser number of constraint tree nodes within the time limit.
This explains why the STT-CBS sometimes does not perform better than CBS in an average of 100 tasks.
The greedy heuristic improves the performance of the ``difficult'' task group by finding a conflict-free solution in more tasks (14 tasks for GSTT-CBS compared to 0 for STT-CBS).
\begin{table}[bt]
    \centering
    \caption{Effect of the greedy heuristic on Map1.}
    {\small
    \begin{tabular}{ll|ccc}
    \hline
        ~ & ~ & GSTT-CBS & STT-CBS & CBS \\ \midrule
        Easy & Num found & 64 & 56 & 66 \\ 
        ~ & Num timeout & 2 & 10 & 0 \\ 
        ~ & Flowtime & 1190.56 & 1191.18 & 1166.62 \\ \hline
        Difficult & Num found & 14 & 0 & 0 \\ 
        ~ & Num timeout & 20 & 34 & 34 \\ 
        ~ & Flowtime & 1778.59 & 1898.78 & 1843.93 \\ \hline
    \end{tabular}
    }
    \label{timeout_analysis}
\end{table}

\subsubsection{Effect of the PU}
\label{analysis_parameter_update}
We conduct an additional experiment to make clear how the parameter of the delay probability distribution is updated using a map with 50 nodes and 1000 tasks consisting of 10 agents. The tasks are solved and simulated sequentially. We compare three cases: (1) the GSTT-CBS with the parameter update; (2) the GSTT-CBS without the parameter update; and (3) the GSTT-CBS with no error, which is a hypothetical setting where the true parameters are obtained from the beginning. After each simulation, the root-mean-square error (RMSE) of the estimated parameter $a_{map}$ and $b_{map}$ of all edges in the map are calculated. The error rate of each edge
 $E_a = \frac{|a - a_{map}|}{|a - a_{prior}|}$ and $ E_b = \frac{|b - b_{map}|}{|b - b_{prior}|}$ is computed as well.
 
\begin{figure}[tb]
  \centering
  \includegraphics[width=1.0\linewidth]{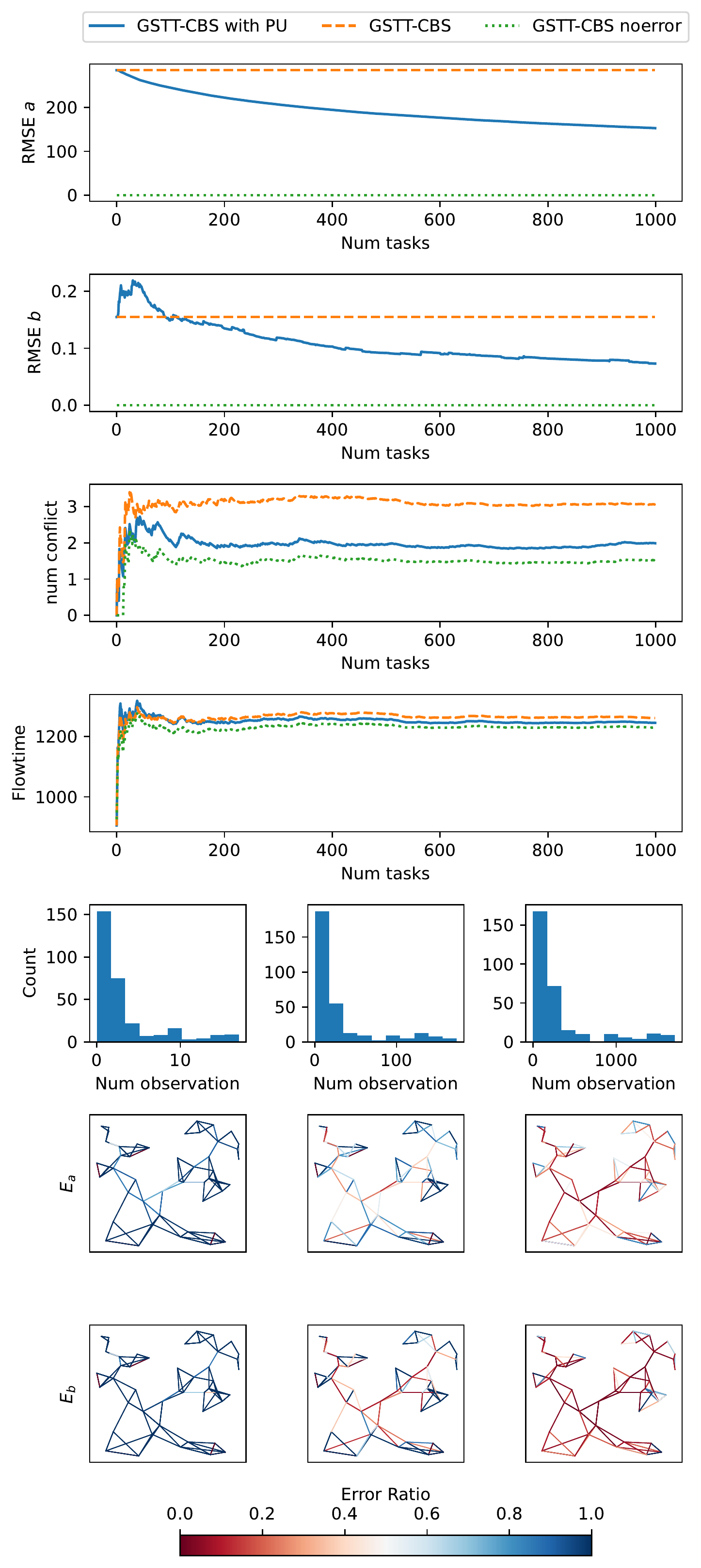}
  \caption{\underline{First and second rows:} RMSE of the estimated $a$ and $b$ values over all edges after each task execution.\\
  \underline{Third and fourth rows:} the number of conflict and flowtime averaged until each execution of tasks.\\
  \underline{Fifth row:} histograms of the number of observations per edge after executing 10 (left), 100 (middle) and 1000 (right) tasks.\\
  \underline{The bottom two rows:} error ratio of each edge plotted on the map, after 10, 100,and 1000 tasks (left to right).
  }
  \label{fig:param_update}
\end{figure}

Figure \ref{fig:param_update} summarizes the analysis results. The RMSE of the estimated value decreases as the number of executed tasks increases. Although the RMSE values do not decrease to 0 even after 1000 tasks, the number of conflicts and flowtime of the GSTT-CBS with PU (shown in the blue line in the third and fourth row) reach their minimum values after about 200 tasks.
The histograms of the number of observations of each edge in the fifth row typically show that only a limited number of edges are frequently used, and many other edges have a small chance of being traversed and their delay being observed.
The bottom two rows of the figure illustrate that the amount of error of the frequently used edges quickly decreases. The estimate on these edges (shown in red) becomes accurate only after 100 iterations. 
Some unimportant edges remain inaccurate, even after 1000 iterations (shown in blue).

The results show that although the RMSE for all edges remains after 1000 trials, the parameters of the edges frequently used in the path planning are updated to be accurate after only 100 to 200 trials.
In this way, the PU can concentrate the effort of data acquisition on important edges, making it possible to create conflict-less solutions after a small number of trials without knowing the true delay probability distributions of all edges.

However, there is still some room for improvement compared to the ideal planning (the gap between blue and green lines in the third and fourth rows). The result indicates that path planning is affected by edges that have not been passed frequently, and exploration for obtaining information on the delay of these edges is necessary. We discuss this point in Section \ref{future_work}.

\subsubsection{Effect of the OR}
We repeat the experiment on Map1 and GSTT-CBS with OR using different values of the time interval of online re-planning $t_{ci} = \{10, 25, 50, 75\}$. This experiment aims to reveal how effective the OR is.
Table \ref{interval_analysis} presents the results. 
The online calculation time in the table is the average of the total calculation time in the re-planning phase (i.e., if the plan is updated 10 times during a task, the total time of the 10 calculations is shown). 
The result shows that a shorter update interval results in fewer conflicts and a shorter flowtime. 
Note that parameter update is not performed; therefore, the planning is based on the delay probability model with the modeling error. 
Even with an inaccurate delay model, frequent re-planning can reduce the number of conflicts.
However, this is at the cost of an increased computation time at OR phase and an increased risk of the calculation results being invalid due to the changes in the agent state during computation.
We set the calculation time limit $t_{limit}=10$ seconds and assume that the agent states do not change during this period. In cases where such an assumption is valid, frequent re-planning can be effective in reducing collisions.

\begin{table}[bt]
    \centering
    \caption{Effect of online re-planning interval on Map1.}
    {\small
    \begin{tabular}{l|cccc}
    \hline
        $t_{ci}$ & 10 & 25 & 50 & 75 \\ \midrule
        Flowtime & 1242.75 & 1264.72 & 1312.24 & 1353.99 \\ 
        Number of Conflict & 0.71 & 1.06 & 1.75 & 2.33 \\ 
        Initial Calculation Time (sec) & 3.02 & 3.03 & 3.01 & 3.20 \\
        Online Calculation Time (sec) & 4.38 & 1.23 & 0.50 & 0.28 \\ \hline
    \end{tabular}
    }
    \label{interval_analysis}
\end{table}

\subsection{Performance on larger instances}
\begin{figure}[tb]
  \centering
  \includegraphics[width=1.0\linewidth]{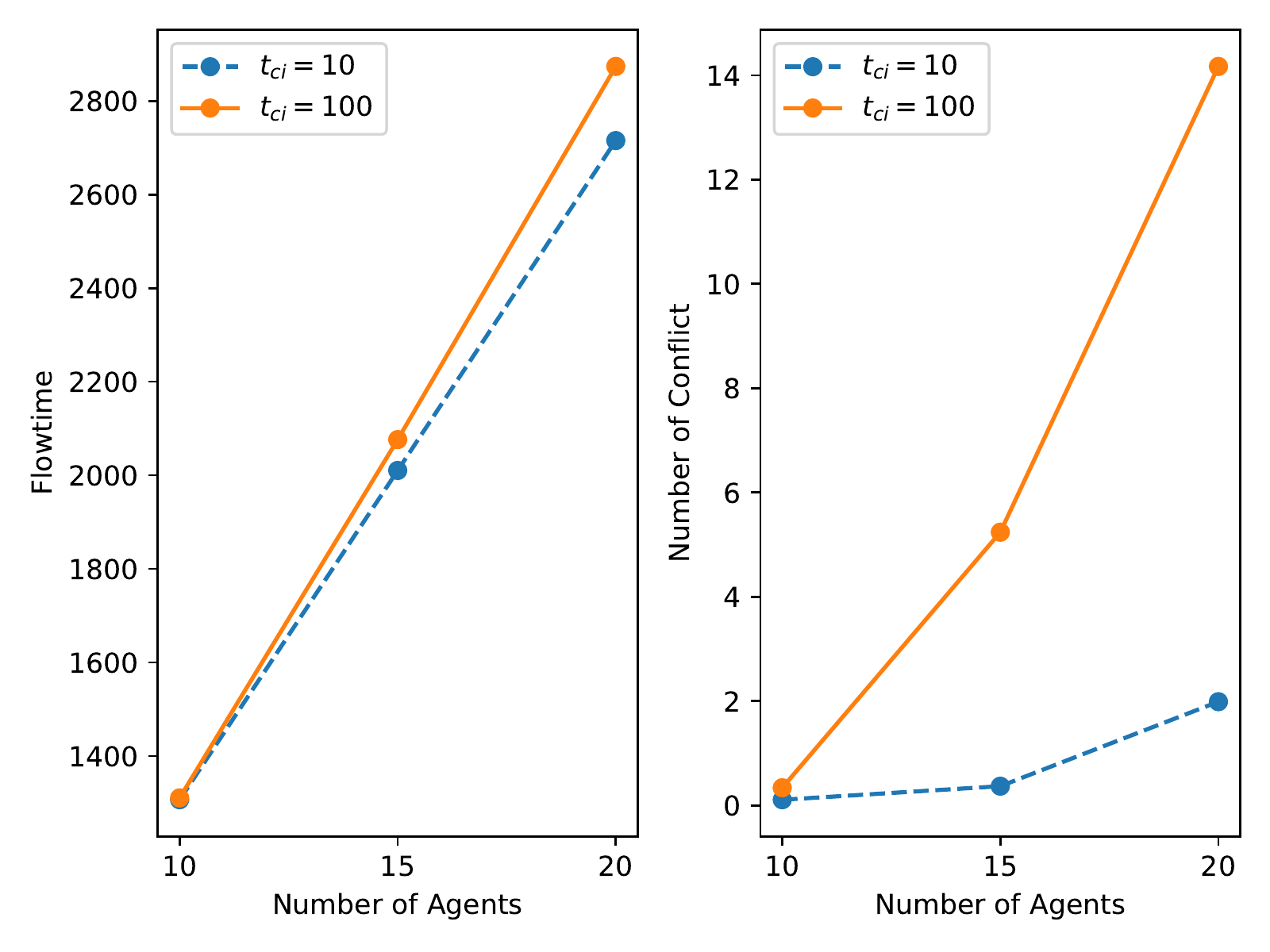}
  \caption{
  Results of the GSTT-CBS with OR and PU on large instances.
  }
  \label{fig:lerge_instance}
\end{figure}

We conduct an additional experiment to answer the question as regards the scalability of our proposed method.
A graph consisting of 100 vertices is generated in the same manner as described earlier. 
A simulation is performed with 100 tasks each consisting of 10, 15, and 20 agents. 
The GSTT-CBS with OR and PU with different values of the constant update interval $t_{ci} = \{10, 100\}$ is tested.
The average flowtime and the average number of conflicts over 100 tasks is depicted in Figure\ref{fig:lerge_instance}.
As the number of agents increases, it becomes more difficult to complete the computation within the time limit. 
More than 60\% of the tasks with 15 agents, and 
more than 90\% of the tasks with 20 agents have reached a timeout. 
However, the frequent re-planning with $t_{ci} = 10$ keeps the number of collisions low.


\section{Conclusion}
In this work, we addressed the MAPF problem with stochastic travel times on non-unit time graphs, 
where the travel time of agents is stochastically distributed (i.e., aleatoric uncertainty), 
but the true distribution is unknown (i.e., epistemic uncertainty) and has to be estimated from the data obtained during the execution of the plan. 
To cope with such a problem, we proposed online re-planning and parameter update methods.
To our knowledge, this is the first study that extends the MAPF problem by introducing these two types of uncertainty to address the real world situations of the automated delivery system in buildings.

We formulated the online MAPF problem, in which the updated plan must be consistent
with currently executed commands, and the solution must be returned within a certain time limit 
to model the real-world situation of automated delivery service. We used a greedy heuristic to find a solution with
less chance of collision in a limited calculation time.

We then compared our method with the existing methods using non-grid graphs with edges that have
a continuous value of travel time and a probabilistic distribution of delay.
The result showed that our proposed method can decrease the number of conflicts
that occur during execution and the sum of the actual travel time of agents.
The performance was achieved with a small number of learning tasks because the parameter update was 
prioritized on frequently used edges.

\section{Future Work}
\label{future_work}
The future work has many possible directions. 

The result of the analysis in \ref{analysis_parameter_update} indicates that
further studies are needed for active exploration methods that use the edges that 
are not in the optimal path in order to obtain information on delay probability of the edge, 
while considering the trade-off between exploration and exploitation. 

In real-world situations where plans must be created in a limited amount of time,
it is effective to prioritize avoidance of timeout over optimality. 
Therefore, adopting other heuristic solvers can be promising. 
Algorithms dedicated to the online re-planning situation
utilizing the result of previous calculations can also be of research interest.
Another possible approach is to avoid collisions during plan execution by adjusting the command execution time, 
instead of re-calculating the plan of all agents.

Updating the plans of a limited number of agents involved in a conflict
expected in the near future under uncertain stochastic travel times
situations can be a future task.

\balance



\bibliographystyle{ACM-Reference-Format} 
\bibliography{sample}


\end{document}